\documentclass[letterpaper, 10 pt, conference]{ieeeconf}  

\IEEEoverridecommandlockouts                              

\usepackage{graphics} 
\usepackage{graphicx}

\usepackage{epsfig} 
\usepackage{amsmath} 
\usepackage{amssymb}  

\usepackage[binary-units]{siunitx}

\usepackage{algorithm}
\usepackage[noend]{algpseudocode} 
\algnewcommand\AAND{\textbf{ and }}
\algnewcommand\Or{\textbf{ or }}

\usepackage{color}
\usepackage{citesort}
\usepackage{url}
\usepackage{breakurl}
\usepackage[breaklinks]{hyperref}

\DeclareMathAlphabet{\pazocal}{OMS}{zplm}{m}{n}

\def \*#1 {mathbf{#1}}
\def \@#1 {\mathbb{#1}}

\DeclareMathAlphabet{\mathpzc}{OT1}{pzc}{m}{it}

\usepackage{array}
\newcolumntype{C}[1]{>{\centering\arraybackslash}p{#1}}
\newcolumntype{M}[1]{>{\raggedright\arraybackslash}p{#1}}

\usepackage{array} 
\newcolumntype{L}[1]{>{\raggedright\let\newline\\\arraybackslash\hspace{0pt}}m{#1}}	
\newcolumntype{S}[1]{>{\centering\let\newline\\\arraybackslash\hspace{0pt}}m{#1}}
\newcolumntype{R}[1]{>{\raggedleft\let\newline\\\arraybackslash\hspace{0pt}}m{#1}}

\usepackage{fancyhdr}
\fancypagestyle{withfooter}{
  
  \fancyfoot[C]{\footnotesize Accepted to the IEEE ICRA Workshop on Field Robotics 2024}
}

\usepackage[nolist,nohyperlinks]{acronym}
\acrodef{fov}[FoV]{Field of View}
\acrodef{tsp}[TSP]{Traveling Salesman Problem}
\acrodef{gvi}[GVI]{General Visual Inspection}
\acrodef{ve}[VE]{Volumetric Exploration}
\acrodef{esdf}[ESDF]{Euclidean Signed Distance Field}
\acrodef{fpso}[FPSO]{Floating Production Storage and Offloading}
\acrodef{iacs}[IACS]{International Association of Classification Societies}
\acrodef{slam}[SLAM]{Simultaneous Localization and Mapping}
\acrodef{bwt}[BWT]{Ballast Water Tank}
\acrodef{ch}[CH]{Cargo Hold}
\acrodef{bwts}[BWTs]{Ballast Water Tanks}
\acrodef{chs}[CHs]{Cargo Holds}
\acrodef{mmWave}[mmWave]{millimeter Wave}

\makeatletter
\renewcommand*{\@opargbegintheorem}[3]{\trivlist
  \item[\hskip \labelsep{\itshape #1\ #2}] \textit{(#3)}\ }
\makeatother

\title{\LARGE \bf
Maritime Vessel Tank Inspection using Aerial Robots: Experience from the field and dataset release
}

\author{Mihir Dharmadhikari, Nikhil Khedekar, Paolo De Petris, Mihir Kulkarni, Morten Nissov and Kostas Alexis
\thanks{This material was supported by the Research Council of Norway under projects SENTIENT (NO-321435) and REDHUS (NO-317773).}
\thanks{The authors are with the Autonomous Robots Lab, Norwegian University of Science and Technology (NTNU), O. S. Bragstads Plass 2D, 7034, Trondheim, Norway {\tt\small mihir.dharmadhikari@ntnu.no}}
}

\begin{document}

\maketitle
\thispagestyle{withfooter}
\pagestyle{withfooter}

\begin{abstract}
This paper presents field results and lessons learned from the deployment of aerial robots inside ship ballast tanks. Vessel tanks including ballast tanks and cargo holds present dark, dusty environments having simultaneously very narrow openings and wide open spaces that create several challenges for autonomous navigation and inspection operations. We present a system for vessel tank inspection using an aerial robot along with its autonomy modules. We show the results of autonomous exploration and visual inspection in $3$ ships spanning across $7$ distinct types of sections of the ballast tanks. Additionally, we comment on the lessons learned from the field and possible directions for future work.
Finally, we release a dataset consisting of the data from these missions along with data collected with a handheld sensor stick.
\end{abstract}

\section{INTRODUCTION}\label{sec:intro}
The shipping industry is a large industry with over $50$K ships operating across the world. Most ships have a long life span and need to undergo regular inspection and maintenance to ensure safe operation. One important component of this inspection is that of the \ac{bwts} and \ac{chs} inside the ships. Both the cargo holds and ballast tanks are prone to corrosion (especially the ballast tank due to exposure to seawater), cracking, and bulking. Currently, the ships need to travel to designated ports for inspection costing them multiple days of idle time resulting in millions of dollars of losses. Furthermore, these tanks are dark, dirty, and dangerous environments with possible pockets of inert gasses inside them, thus rendering them extremely hazardous workplaces. 

Motivated by the above, a niche community has investigated the use of robotic systems to replace or assist in the inspection process to keep humans out of the way~\cite{knukkel2023remote,krystosik2021use}. These include works on aerial and crawling robots~\cite{carrara2020robotics}, solutions revolving around manually piloted robots~\cite{muhammad2022maritime}, techniques for detecting and navigating through the narrow openings in the \ac{bwts} called manholes~\cite{andersen2022depth,manhole2023}, as well as rail-guided robots~\cite{rijnbeek2015rail,tieleman2016rail,christensen2011tank}, and underwater Remotely Operated Vehicles (ROV)~\cite{thongpool2015application,andritsos2003rotis}. 

\begin{figure}[h!]
\centering
    \includegraphics[width=0.99\columnwidth]{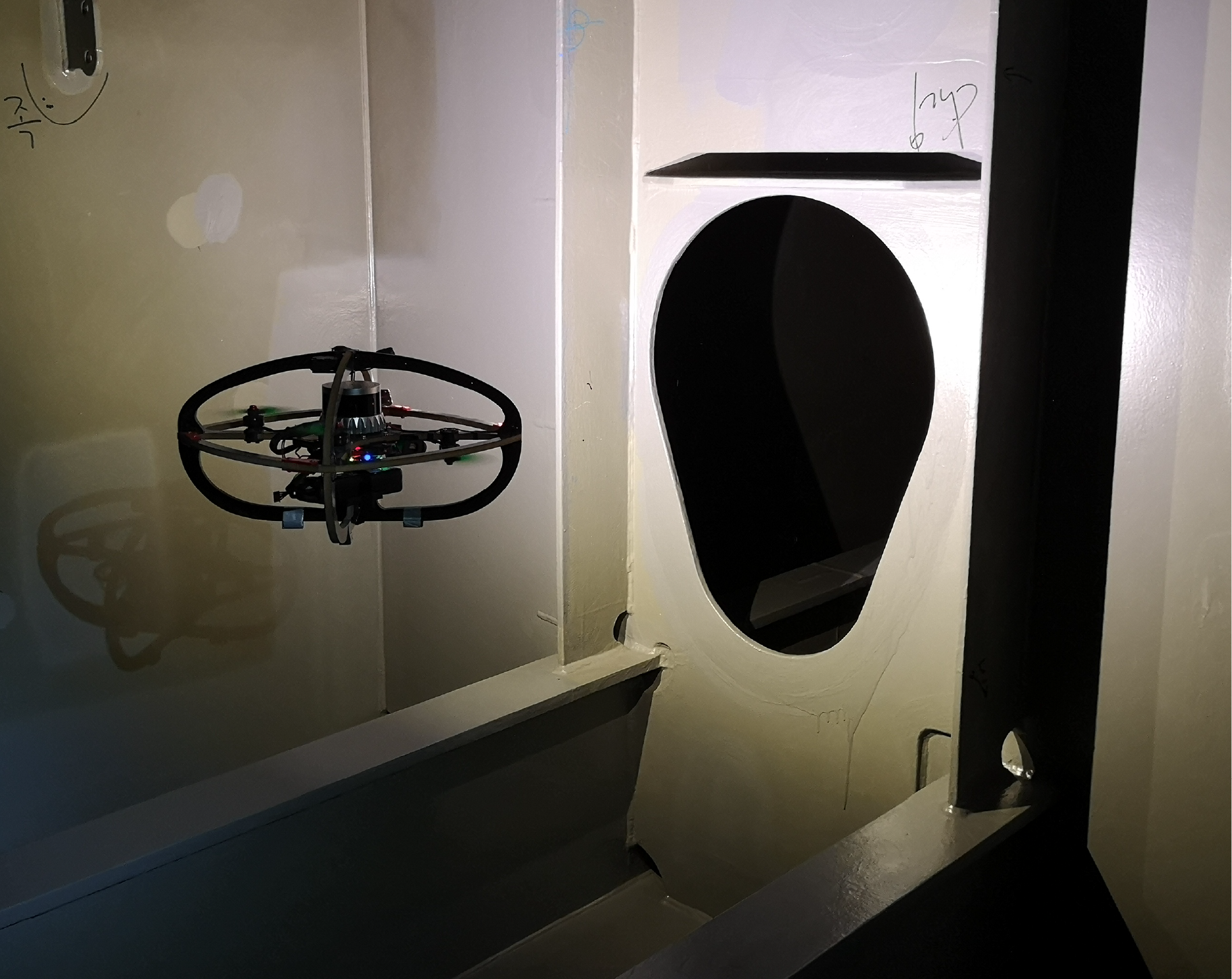}
\caption{Instance of RMF-Owl autonomously operating in a double bottom section of a ballast tank inside a Floating Production Storage and Offloading (FPSO) vessel. }
\label{fig:intro_fig}
\vspace{-3ex} 
\end{figure}

The majority of the solutions utilize limited autonomy, and a smaller portion of it is field-tested in a diverse set of environments. Motivated by the above, our prior work~\cite{2023expgvi} presents a method for fully autonomous exploration and general visual inspection of multiple compartments of ballast tanks that is demonstrated in three field experiments. This paper provides an extensive description of all the field trials conducted during and beyond~\cite{2023expgvi}. We present data from $3$ ships across $7$ different types of sections of ballast tanks collected using RMF-Owl~\cite{rmfowl}, a collision-tolerant aerial robot, during autonomous exploration and inspection missions, manual flights spanning across multiple levels, as well as from a handheld sensor setup, called Mjolnir, consisting of LiDAR, cameras, \ac{mmWave} radars, time of flights sensors, and IMUs. All the data collected from the field deployments is released publicly at \url{https://github.com/ntnu-arl/ballast\_water\_tank\_dataset}.

The remainder of this paper is organized as follows. Section~\ref{sec:system} explains the robotic system and the autonomy modules running onboard, section~\ref{sec:deployment} details the field deployments and released dataset, followed by the lessons learned discussed in section~\ref{sec:lessons}, and conclusions drawn in section~\ref{sec:concl}.  


\section{SYSTEM DESCRIPTION}\label{sec:system}
To inspect the challenging and diverse ballast tanks and cargo holds in ships we used our in-house developed collision-tolerant aerial robot called RMF-Owl~\cite{rmfowl}  that was further modified for improved compute and sensing. The robot design and the components of the onboard autonomy stack are described in the following subsections.

\subsection{Robot Design}
RMF-Owl is designed for fully autonomous operation, lightweight, and collision-tolerance. The main frame of the robot is made from sheets of carbon-foam sandwich material. This enables having a lightweight airframe capable of withstanding the collisions that are bound to happen in the confined areas of the ballast tanks. The sensor suite of the robot consists of an Ouster OS0-64 LiDAR, a Flir Blackfly S 0.4MP color camera, and a VectorNav VN100 IMU. The entire autonomy stack consisting of SLAM, Path planning, and control runs on the onboard Khadas VIM4 Single Board Computer. Finally, we use a PixRacer Pro as the low-level autopilot for thrust and attitude control. All components of the robot can be seen in Figure~\ref{fig:sensor_setup}. With all the payload the robot weights $\SI{1.45}{kg}$ with a flight time of $\SI{10}{minutes}$. Unlike the control approach described in~\cite{rmfowl}, we use a linear model predictive controller~\cite{mpc_rosbookchapter} for position control that provides the roll, pitch, yaw rate, and thrust commands, which are then followed by the attitude controller of the autopilot. Further details about the robot design can be found in~\cite{rmfowl}.

\subsection{Onboard Localization and Mapping}
For accurate localization and mapping, we utilize CompSLAM~\cite{khattak2020complementary}, a method for multi-modal \ac{slam} that combines LiDAR, vision, and IMU data in a hierarchical fashion. Under the hood, the result of a visual-inertial estimator \cite{bloesch2015robust} is health-checked based on a D-Optimality criterion and used as a prior for LiDAR-Inertial registration. The prior is passed through when the LiDAR Inertial registration fails its own health check which relies on thresholding the eigenvalues of the approximate hessian\cite{zhang-on-degen}. This design can handle cases of degeneracy and sensor degradation for a modality in the pipeline providing a robust estimate of the robot pose.

\subsection{Exploration and Inspection Path Planning}
We utilize the autonomous exploration and inspection method described in~\cite{2023expgvi} - which is built on top of our open sourced Graph-based exploration path planner (GBPlanner)~\cite{GBPLANNER_JFR_2020,GBPLANNER2COHORT_ICRA_2022} - onboard the robot. The method is primarily designed for inspection of multiple compartments of a ballast tank but its extension to a cargo hold is trivial as it can be considered as one large compartment. The planner does not assume access to a prior map but only uses rough locations and dimensions of the compartments, a piece of information that is readily available through 2D layouts. Each compartment is tackled separately by first performing volumetric exploration to generate the map of the compartment using LiDAR and then calculating an inspection path to view all mapped surfaces using a camera. To navigate between compartments connected by manholes, the strategy presented in~\cite{manhole2023} is used. Further details and results about the path planning pipeline can be found in~\cite{2023expgvi}.

\section{FIELD DEPLOYMENT}\label{sec:deployment}
\begin{figure}[h!]
\centering
    \includegraphics[width=0.99\columnwidth]{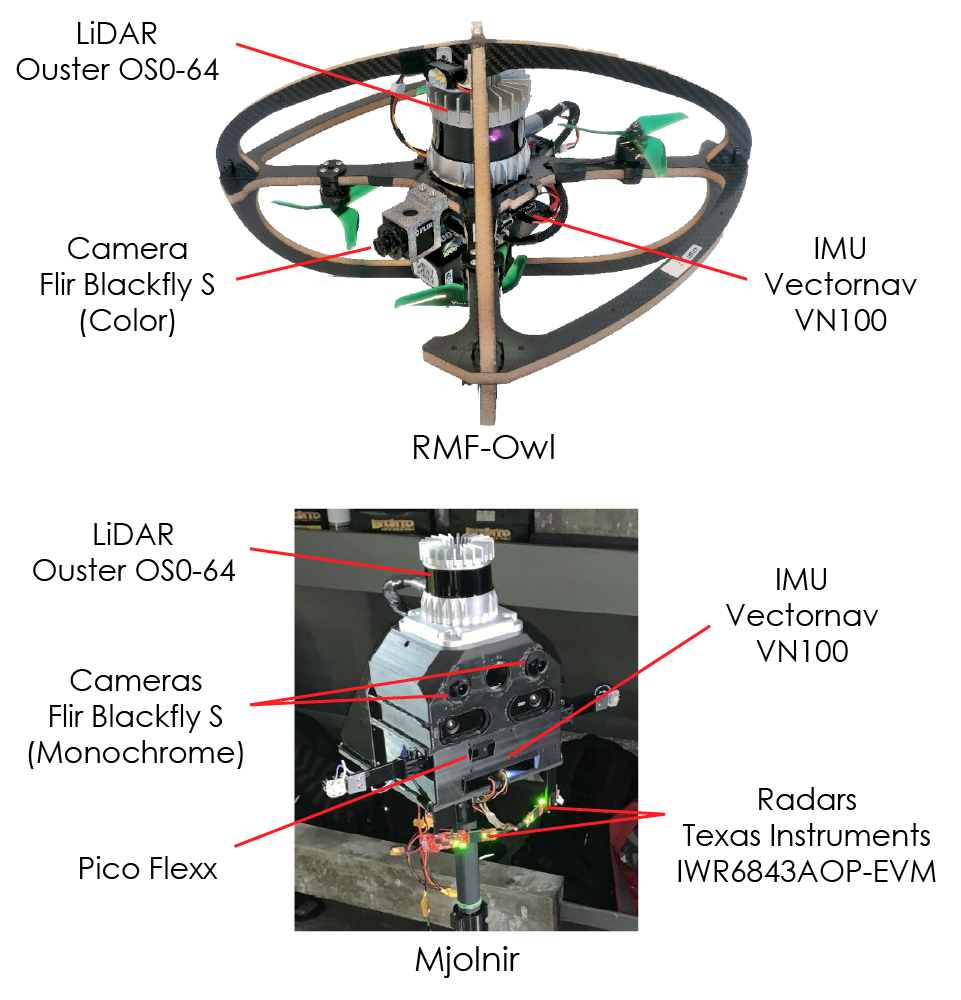}
\caption{Illustration of the sensing and computing on the RMF-Owl aerial robot and Mjolnir handheld sensor stick. All the autonomy modules including SLAM, Path planning, and Control run completely onboard the robot.}
\label{fig:sensor_setup}
\vspace{-3ex} 
\end{figure}

\begin{figure*}[h!]
\centering
    \includegraphics[width=0.99\textwidth]{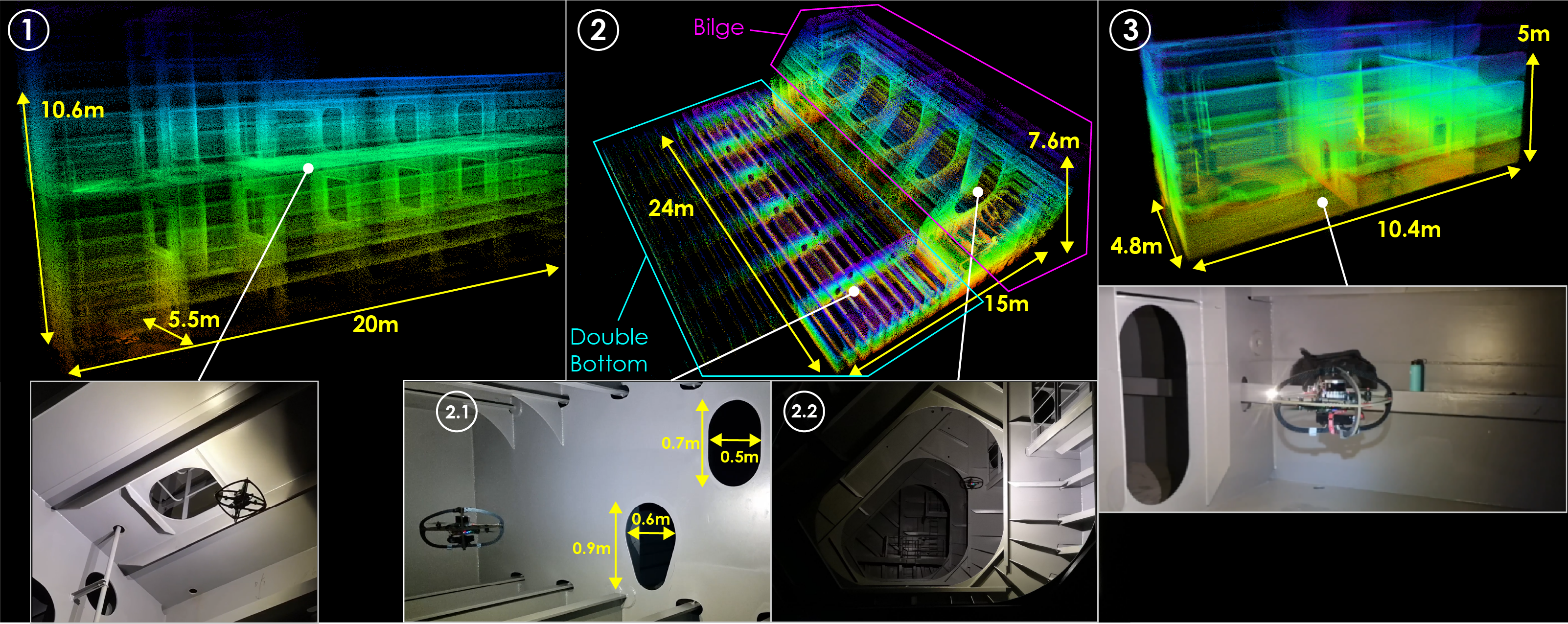} 
\caption{Final maps, as generated onboard the robot, along with instances of deployment of the RMF-Owl aerial robot in FPSO2. The robot was deployed in the side sections (1),(3), double bottom (d.b.) section (2.1), and bilge (2.2) section. We present both autonomous exploration-inspection missions and manual flights. The dataset covers missions in vastly different sections, navigation through extremely narrow manholes ($\SI{0.7}{\meter} \times \SI{0.5}{\meter}$), and multi-level missions.}
\label{fig:fpso2}
\vspace{-3ex} 
\end{figure*}

We conducted field deployments across $3$ vessels in $7$ distinct sections of different types of ballast tanks. An overview of the different environments can be seen in Figures~\ref{fig:fpso1},~\ref{fig:fpso2},~\ref{fig:ot}. To maintain anonymity, we will refer to the ships as FPSO1, FPSO2, and the Oil Tanker (OT). 
In all three deployments, the RMF-Owl collision-tolerant robot was used for autonomous and manual flights. Additionally, a handheld sensor setup, called Mjolnir, was used for data collection during the deployment in FPSO1. The sensor suite on it can be seen in Figure~\ref{fig:sensor_setup}. All the data collection missions are summarised in Table~\ref{tab:test_summary}.
We now describe the environments and deployments in each vessel in detail.

\subsection{FPSO1}
\begin{figure}[h!]
\centering
    \includegraphics[width=0.99\columnwidth]{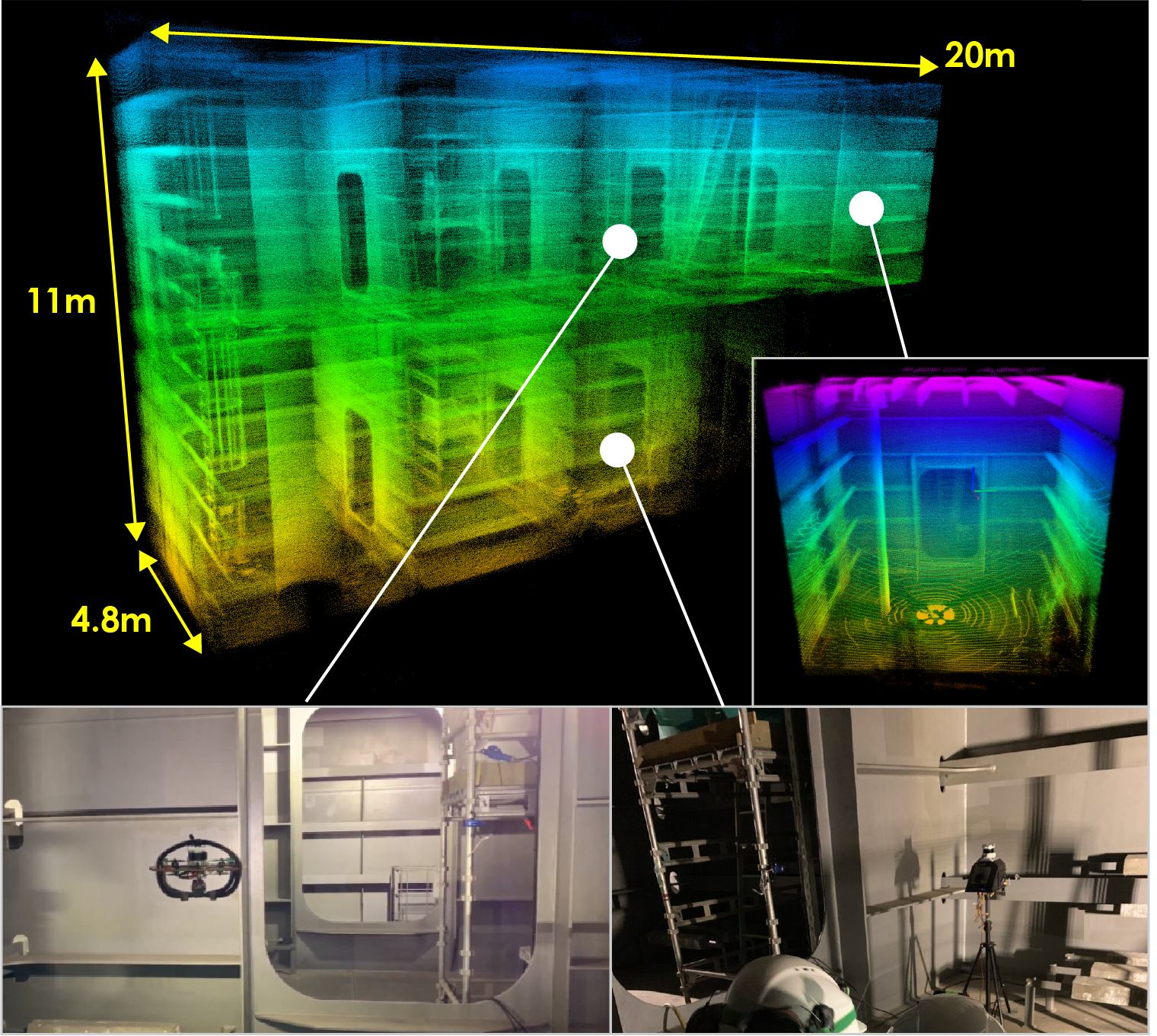} 
\caption{Final maps, as generated onboard the robot, along with instances of the RMF-Owl aerial robot performing exploration and inspection mission and data collection using Mjolnir in FPSO1. We conducted a total of $3$ autonomous and $1$ manual flights using RMF-Owl and $4$ handheld data collection missions using Mjolnir.}
\label{fig:fpso1}
\vspace{-3ex} 
\end{figure}

The experiments in FPSO1 involved the deployment in two levels on a side ballast tank. Both levels consisted of $5$ compartments connected by large openings. All compartments were cubical in shape and identical in terms of dimensions and the majority of structural components. The dimensions of the tank and some images of the interior can be seen in Figure~\ref{fig:fpso1}. We conducted three autonomous exploration and general visual inspection missions where the robot was tasked to inspect $1, 3$, and $5$ compartments respectively on level 1 of the tank. The maximum allowed height was restricted in the last two missions. Additionally, a manual flight was conducted to collect data across both levels. The robot was flown manually on each level and pulled up from level 2 to level 1 using a rope. A rope was used as an upward-facing camera for piloting the robot through the access hatch was not integrated on the robot at that time. The final maps are shown in Figure~\ref{fig:fpso1}. It is noted that all maps shown in the paper are presented as they were generated onboard the robot.

\subsection{FPSO2}
The deployment in FPSO2 spanned across $4$ different sections of the ballast tank namely, side tank sections - with and without manholes - (Figure~\ref{fig:fpso2}.1, \ref{fig:fpso2}.3), the double bottom section (Figure~\ref{fig:fpso2}.2.1), and the bilge section (Figure~\ref{fig:fpso2}.2.2). The compartments in the side sections were cubical in shape with either large openings to pass through or with manholes of dimensions $\SI{1.3}{\meter} \times \SI{0.6}{\meter}$. The bilge section is the last level on the side of the ship connecting the side sections with the double bottom sections. This area is much wider compared to the other sections (dimensions shown in Figure~\ref{fig:fpso2}). It is divided into multiple compartments. They have manholes connecting them at a lower height for humans to go through, but there are also larger openings at the top. Finally, the double bottom section spans across the bottom of the ship, it is much smaller in height but wider and has multiple extremely narrow manholes (smallest manhole being $\SI{0.7}{\meter} \times \SI{0.5}{\meter}$ in dimension).
Autonomous exploration and inspection missions were conducted in the first three sections as well as an exploration-only mission in the double-bottom section requiring the robot to pass through the narrowest manholes. Additionally, we collected a larger dataset through two manual flights. The first covers two levels of the side tank section with the robot being piloted through the access hatch across the levels (Figure~\ref{fig:fpso2}.1). The second flight covers the bilge and the double bottom section (Figure~\ref{fig:fpso2}.2). 

\begin{figure}[h!]
\centering
    \includegraphics[width=0.99\columnwidth]{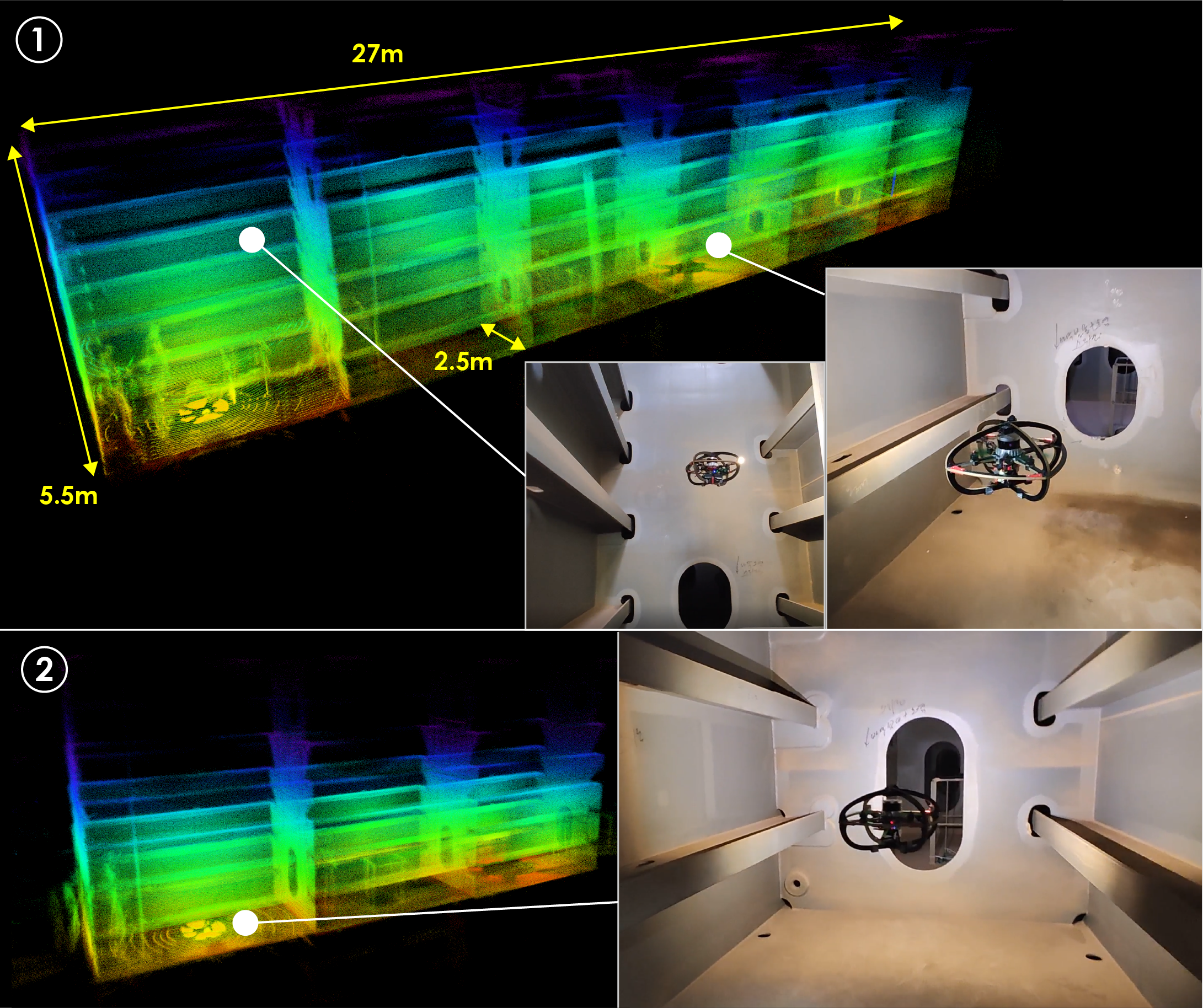} 
\caption{Final maps, as generated onboard the robot, along with instances of the RMF-Owl aerial robot performing exploration and inspection missions in the Oil Tanker. All $5$ tests were conducted in the side tanks covering $3$ - $6$ compartments. The environment contains two types of manholes $\SI{0.8}{\meter} \times \SI{0.6}{\meter}$ and $\SI{1.2}{\meter} \times \SI{0.6}{\meter}$.}
\label{fig:ot}
\vspace{-3ex} 
\end{figure}

\subsection{Oil Tanker (OT)}
All the tests in the OT were conducted in the side tank sections. The compartments in this ballast tank were narrower ($\SI{2.5}{\meter}$) than the other two vessels ($\SI{4.8}{\meter}$, $\SI{5.5}{\meter}$). The missions were conducted on two levels with cubical-shaped compartments connected by manholes of dimensions $\SI{0.8}{\meter} \times \SI{0.6}{\meter}$ on level 2 (Figure~\ref{fig:ot}.1) or $\SI{1.2}{\meter} \times \SI{0.6}{\meter}$ on level 1 (Figure~\ref{fig:ot}.2). A total of $5$ missions were conducted, $3$ on level 1 and $2$ on level 2. The maps from one mission on each level are shown in Figure~\ref{fig:ot} along with instances of the robot in the tank to showcase the environment. 

\begin{table}[]
\centering
\vspace{-1ex}
\caption{Summary of all Field deployments}
\begin{tabular}{|l|l|l|l|l|l|}
\hline
\textbf{No.} & \textbf{Ship} & \textbf{Section} & \textbf{Autonomous} & \textbf{Multi-level} & \textbf{Time(s)}           \\ \hline
\multicolumn{6}{|c|}{\textbf{RMF-Owl}} \\ \hline
1  & FPSO1 & side & Yes & No & 225 \\ \hline
2  & FPSO1 & side & Yes & No & 300 \\ \hline
3  & FPSO1 & side & Yes & No & 154 \\ \hline
4  & FPSO1 & side & No & Yes & 450 \\ \hline
5  & FPSO2 & side & Yes & No & 300 \\ \hline
6  & FPSO2 & side & Yes & No & 380 \\ \hline
7  & FPSO2 & bilge & Yes & No & 336 \\ \hline
8  & FPSO2 & d.b. & Yes & No & 200 \\ \hline
9  & FPSO2 & side & No & Yes & 258 \\ \hline
10 & FPSO2 & bilge, d.b. & No & No & 275 \\ \hline
11 & OT & side    & Yes & No & 214 \\ \hline
12 & OT & side    & Yes & No & 216 \\ \hline
13 & OT & side    & Yes & No & 354 \\ \hline
14 & OT & side    & Yes & No & 360 \\ \hline
15 & OT & side    & Yes & No & 370 \\ \hline
\multicolumn{6}{|c|}{\textbf{Mjolnir}} \\ \hline
1  & FPSO1 & side & N/A & No & 268 \\ \hline
2  & FPSO1 & side & N/A & No & 373 \\ \hline
3  & FPSO1 & side & N/A & No & 357 \\ \hline
4  & FPSO1 & side & N/A & No & 395 \\ \hline
\end{tabular}
\label{tab:test_summary}
\end{table}

The data collected from the above deployments is made publicly available at \url{https://github.com/ntnu-arl/ballast\_water\_tank\_dataset}.

\section{LESSONS LEARNED}\label{sec:lessons}

\subsection{Resilient Autonomy:}
\begin{itemize}
    \item \textbf{Collision-tolerant robot design}: As the robot needs to be able to navigate through narrow manholes with very small clearances, a scenario where the robot collides with the environment is inevitable. Additionally, any drift in odometry can increase these chanced further. Hence, collision-tolerant robot design is key for resilient navigation in these environments.
    \item \textbf{Robust and High Precision Mapping}: Precise localization of defects across multiple inspection missions is critical for tracking the health of the tanks. Hence, robust and accurate localization and mapping is key for tank inspections. One of the important defects that needs to be detected early on is the deformation of load-bearing structures. This requires the generation of high precision and high resolution (order or $\SI{1}{cm}$) $3$D maps which can be used to identify such deformations.
    \item \textbf{Scalability}: Vessel tanks present a diverse set of environments containing narrow openings like the manholes as well as extremely wide open spaces like cargo holds or the bilge section of ballast tanks. Furthermore, the autonomy module needs to optimize different types of objectives (e.g., volumetric exploration and mapping, general visual inspection, closeup visual inspection, manhole passing, etc.). This requires the autonomy solution to be capable of providing a set of behaviors that are triggered based on the scenario and mission objectives.
\end{itemize}

\subsection{Semantic Reasoning:}
\begin{itemize}
    \item Vessel tanks contain certain structures and objects that are more prone to fatigue/stress and hence are more important for inspection than the rest of the surfaces. We refer to them as inspection semantics. Detection, localization, and building a representation of these semantics for planning are important to increase the efficiency and effectiveness of the inspection.
    \item These semantics are more susceptible to defects and hence require inspection from a closer distance. If all inspection-important and unimportant areas are treated equally and inspected with the same quality, the robot won't be able to undertake large-scale missions spanning multiple levels of the ballast tank. Hence, it is necessary to identify and focus on the inspection-important semantics.
    \item The spatial arrangement of these semantics repeats across compartments on one level of the ballast tank, and their general topology also has distinctive characteristics that repeat across tanks. Such structural patterns can be exploited in the planning pipeline to take more informed actions.
\end{itemize}

\subsection{Image quality for defect detection:}
\begin{itemize}
    \item Majority of the defects such as corrosion, cracks, etc. are detected on camera images, hence, good quality images are key to good inspection. Image quality depends on the optics, image resolution, and light conditions. Miniaturization of the first is hard to achieve and can limit the low-light capabilities of the camera.
    \item Creating optimal lighting conditions is a non-trivial task due to the nature of the defects. Due to the lack of external light inside the tank, the robot needs to carry onboard lighting. Detection of larger defects such as corrosion does not require any special lighting conditions. However, certain hairline cracks are not visible if the light is shining perpendicular to the surface. Human inspectors generally shine the light at an angle to the surface which highlights such cracks better. Such active light adjustment behavior needs to be incorporated in the autonomy as well to be able to detect such cracks. 
\end{itemize}

\section{CONCLUSIONS}\label{sec:concl}
In this paper, we present an aerial robotic system for the inspection of maritime vessel tanks along with its onboard autonomy modules. The system is field evaluated in $3$ ships and the details of the deployment are presented. A dataset collected from these deployments using the proposed robotic system as well as a handheld sensor stick is released publicly. The key lessons learned from these field tests are presented for the research community to build upon.

\section{ACKNOWLEDGEMENTS}
We would like to thank our industry partners in the SENTIENT and REDHUS projects. These two projects are funded by the Research Council of Norway and involve industry collaboration. SENTIENT is led by NTNU, whereas REDHUS is led by DNV.

\addtolength{\textheight}{-12cm}   




\bibliographystyle{IEEEtran}
\bibliography{./BWTGVI_ICAR_2023}

\begin{thebibliography}{10}
\providecommand{\url}[1]{#1}
\csname url@samestyle\endcsname
\providecommand{\newblock}{\relax}
\providecommand{\bibinfo}[2]{#2}
\providecommand{\BIBentrySTDinterwordspacing}{\spaceskip=0pt\relax}
\providecommand{\BIBentryALTinterwordstretchfactor}{4}
\providecommand{\BIBentryALTinterwordspacing}{\spaceskip=\fontdimen2\font plus
\BIBentryALTinterwordstretchfactor\fontdimen3\font minus \fontdimen4\font\relax}
\providecommand{\BIBforeignlanguage}[2]{{%
\expandafter\ifx\csname l@#1\endcsname\relax
\typeout{** WARNING: IEEEtran.bst: No hyphenation pattern has been}%
\typeout{** loaded for the language `#1'. Using the pattern for}%
\typeout{** the default language instead.}%
\else
\language=\csname l@#1\endcsname
\fi
#2}}
\providecommand{\BIBdecl}{\relax}
\BIBdecl

\bibitem{knukkel2023remote}
D.~Knukkel, ``Remote inspection schemes: Past, present, and future,'' in \emph{Smart Ports and Robotic Systems: Navigating the Waves of Techno-Regulation and Governance}.\hskip 1em plus 0.5em minus 0.4em\relax Springer, 2023, pp. 327--342.

\bibitem{krystosik2021use}
A.~Krystosik-Gromadzi{\'n}ska, ``The use of drones in the maritime sector--areas and benefits,'' \emph{Zeszyty Naukowe Akademii Morskiej w Szczecinie}, 2021.

\bibitem{carrara2020robotics}
E.~Carrara and A.~Grasso, ``Robotics technology for inspection of ships,'' in \emph{2020 25th IEEE International Conference on Emerging Technologies and Factory Automation (ETFA)}, vol.~1.\hskip 1em plus 0.5em minus 0.4em\relax IEEE, 2020, pp. 1526--1533.

\bibitem{muhammad2022maritime}
B.~Muhammad and A.~Gregersen, ``Maritime drone services ecosystem-potentials and challenges,'' in \emph{2022 IEEE International Black Sea Conference on Communications and Networking (BlackSeaCom)}.\hskip 1em plus 0.5em minus 0.4em\relax IEEE, 2022, pp. 6--13.

\bibitem{andersen2022depth}
R.~E. Andersen, M.~Zajaczkowski, H.~Jaiswal, J.~Xu, W.~Fan, and E.~Boukas, ``Depth-based deep learning for manhole detection in uav navigation,'' in \emph{2022 IEEE International Conference on Imaging Systems and Techniques (IST)}.\hskip 1em plus 0.5em minus 0.4em\relax IEEE, 2022, pp. 1--6.

\bibitem{manhole2023}
M.~Dharmadhikari, P.~De~Petris, H.~Nguyen, M.~Kulkarni, N.~Khedekar, and K.~Alexis, ``Manhole detection and traversal for exploration of ballast water tanks using micro aerial vehicles,'' in \emph{2023 International Conference on Unmanned Aircraft Systems (ICUAS)}, 2023, pp. 103--109.

\bibitem{rijnbeek2015rail}
E.~Rijnbeek, ``Rail-guided ballast tank inspection robot, hardware infrastructure analysis and enhancement,'' {B.S.} thesis, University of Twente, 2015.

\bibitem{tieleman2016rail}
P.~Tieleman, ``Rail-guided ballast tank inspection robot, software architecture analysis and enhancement in the ros environment,'' {B.S.} thesis, University of Twente, 2016.

\bibitem{christensen2011tank}
L.~Christensen, J.~Lemburg, T.~V{\"o}gele, F.~Kirchner, N.~Fischer, R.~Ahlers, G.~Psarros, and L.-E. Etzold, ``Tank inspection by cost effective rail based robots,'' in \emph{ICCAS}.\hskip 1em plus 0.5em minus 0.4em\relax Trieste, Italy, 2011.

\bibitem{thongpool2015application}
K.~Thongpool, T.~Preeyanurak, A.~Poonpon, and P.~Insuk, ``Application of mini-rov technology for fso ballast tank inspection and thickness gauging,'' in \emph{SPE/IATMI Asia Pacific Oil \& Gas Conference and Exhibition}.\hskip 1em plus 0.5em minus 0.4em\relax OnePetro, 2015.

\bibitem{andritsos2003rotis}
F.~Andritsos and D.~Maddalena, ``Rotis: Remotely operated tanker inspection system,'' in \emph{Proceedings of the 8th International Marine Design Conference (lMDC), Athens, Greece}, 2003, pp. 5--8.

\bibitem{2023expgvi}
M.~Dharmadhikari, P.~De~Petris, M.~Kulkarni, N.~Khedekar, H.~Nguyen, A.~E. Stene, E.~Sjøvold, K.~Solheim, B.~Gussiaas, and K.~Alexis, ``Autonomous exploration and general visual inspection of ship ballast water tanks using aerial robots,'' in \emph{2023 21st International Conference on Advanced Robotics (ICAR)}, 2023, pp. 409--416.

\bibitem{rmfowl}
P.~D. Petris, H.~Nguyen, M.~Dharmadhikari, M.~Kulkarni, N.~Khedekar, F.~Mascarich, and K.~Alexis, ``Rmf-owl: A collision-tolerant flying robot for autonomous subterranean exploration,'' in \emph{2022 International Conference on Unmanned Aircraft Systems (ICUAS)}, 2022, pp. 536--543.

\bibitem{mpc_rosbookchapter}
{M. Kamel, T. Stastny, K. Alexis, and R. Siegwart}, ``Model predictive control for trajectory tracking of unmanned aerial vehicles using ros,'' \emph{Springer Book on Robot Operating System (ROS)}.

\bibitem{khattak2020complementary}
S.~Khattak, H.~Nguyen, F.~Mascarich, T.~Dang, and K.~Alexis, ``Complementary multi--modal sensor fusion for resilient robot pose estimation in subterranean environments,'' in \emph{2020 International Conference on Unmanned Aircraft Systems (ICUAS)}.\hskip 1em plus 0.5em minus 0.4em\relax IEEE, 2020, pp. 1024--1029.

\bibitem{bloesch2015robust}
M.~Bloesch, S.~Omari, M.~Hutter, and R.~Siegwart, ``Robust visual inertial odometry using a direct ekf-based approach,'' in \emph{Intelligent Robots and Systems (IROS), 2015 IEEE/RSJ International Conference on}.\hskip 1em plus 0.5em minus 0.4em\relax IEEE, 2015, pp. 298--304.

\bibitem{zhang-on-degen}
J.~Zhang, M.~Kaess, and S.~Singh, ``On degeneracy of optimization-based state estimation problems,'' in \emph{2016 IEEE International Conference on Robotics and Automation (ICRA)}, 2016, pp. 809--816.

\bibitem{GBPLANNER_JFR_2020}
T.~Dang, M.~Tranzatto, S.~Khattak, F.~Mascarich, K.~Alexis, and M.~Hutter, ``Graph-based subterranean exploration path planning using aerial and legged robots,'' \emph{Journal of Field Robotics}, vol.~37, no.~8, pp. 1363--1388, 2020.

\bibitem{GBPLANNER2COHORT_ICRA_2022}
M.~Kulkarni, M.~Dharmadhikari, M.~Tranzatto, S.~Zimmermann, V.~Reijgwart, P.~De~Petris, H.~Nguyen, N.~Khedekar, C.~Papachristos, L.~Ott, R.~Siegwart, M.~Hutter, and K.~Alexis, ``Autonomous teamed exploration of subterranean environments using legged and aerial robots,'' in \emph{2022 International Conference on Robotics and Automation (ICRA)}.\hskip 1em plus 0.5em minus 0.4em\relax IEEE, 2022, pp. 3306--3313.

\end{thebibliography}

\end{document}